\documentclass[1p,review]{elsarticle}
\usepackage{hyperref}
\usepackage[utf8]{inputenc} % allow utf-8 input
\usepackage[T1]{fontenc}    % use 8-bit T1 fonts
\usepackage[tbtags,fixamsmath]{mathtools}
\usepackage{subfigure}
\usepackage{booktabs,multirow,multicol,ctable}
\usepackage{amsmath,amssymb} % define this before the line numbering.
\usepackage{dsfont}

\def\HangBox#1{%^
\begin{minipage}[t]{\textwidth}% Top-hanging minipage, will align on
\begin{tabbing} % tabbing so that minipage shrinks to fit
~\\[-\baselineskip] % Make first line zero-height
#1 % Include user's text
\end{tabbing}%^
\end{minipage}} % can't allow } onto next line, as {WIDEBOX}~x will not tie.

\usepackage{color}
\usepackage[normalem]{ulem}

\usepackage{xcolor}
\usepackage{soul}
\usepackage[normalem]{ulem}

\def\mypar#1{\vspace{0.15cm}\noindent{\bf #1.}}

\def\rev#1{{#1}} % red

\IfFileExists{darkmode.tex}{
    \usepackage{pagecolor}
    \definecolor{myfg}{gray}{0.94} 
    \definecolor{mybg}{gray}{0}
    \input{darkmode.tex} % myfg and mybg can be altered in darkmode.tex
    \pagecolor{mybg}
    \color{myfg}
}{} 

\usepackage{amssymb} % --> nicer

\def\etal{et al.\ }

\usepackage{amsmath,amssymb,amsbsy,xspace}
\makeatletter
\DeclareRobustCommand\onedot{\futurelet\@let@token\@onedot}
\def\@onedot{\ifx\@let@token.\else.\null\fi\xspace}

\def\eg{\emph{e.g}\onedot}

\def\etal{\emph{et al}\onedot}

\makeatother

\journal{Image and Vision Computing (IMAVIS)}

\begin{document}

\begin{frontmatter}

\title{Caption Generation on Scenes with Seen and Unseen Object Categories}

\author[odtuaddress,havelsanaddress]{Berkan Demirel}
\ead{berkan.demirel@metu.edu.tr}

\author[odtuaddress]{Ramazan Gokberk Cinbis}
\ead{gcinbis@ceng.metu.edu.tr}

\address[odtuaddress]{Department of Computer Engineering, Middle East Technical University, 06800 Ankara, Turkey}
\address[havelsanaddress]{Image and Video Processing Group, HAVELSAN Inc., 06800 Ankara, Turkey}

\begin{abstract}
Image caption generation is one of the most challenging problems at the intersection of vision and language domains. In this work, we propose a realistic captioning task where the input scenes may incorporate visual objects with no corresponding visual or textual training examples. For this problem, we propose a detection-driven approach that consists of a single-stage generalized zero-shot detection model to recognize and localize instances of both seen and unseen classes, and a template-based captioning model that transforms detections into sentences. To improve the generalized zero-shot detection model, which provides essential information for captioning, we define effective class representations in terms of class-to-class semantic similarities, and leverage their special structure to construct an effective unseen/seen class confidence score calibration mechanism. We also propose a novel evaluation metric that provides additional insights for the captioning outputs by separately measuring the visual and non-visual contents of generated sentences. Our experiments highlight the importance of studying captioning in the proposed zero-shot setting, and verify the effectiveness of the proposed detection-driven
zero-shot captioning approach.
\end{abstract}


\end{frontmatter}

\section{Introduction}
\label{sec:intro}

The problem of generating a concise textual summary of a given image, known as {\em image captioning}, is one of the most challenging
problems that require joint vision and lingual modeling. With ever-increasing recognition rates in object detection models, 
pioneered by \cite{redmon2016yolo9000, redmon2016you, girshick2015fast,
sermanet2014overfeat, bell2016inside, ren2015faster, lin2017feature, liu2016ssd, yan2014fastest,lin2018focal,
law2018cornernet}, there has been a recent interest in generating visually grounded captions via constructing
detection-driven captioning models, \eg \cite{kulkarni2013babytalk, anne2016deep,
yin2017obj2text, lu2018neural}. However, the success of such approaches is inherently limited by the set of
classes spanned by the detector training set, which is typically too small to construct a visually comprehensive model. Therefore, such models are prone to 
synthesizing irrelevant captions in realistic, uncontrolled settings where input images may contain instances of 
classes unseen during training.

\begin{figure}
\begin{center}
\includegraphics[width=0.6\textwidth]{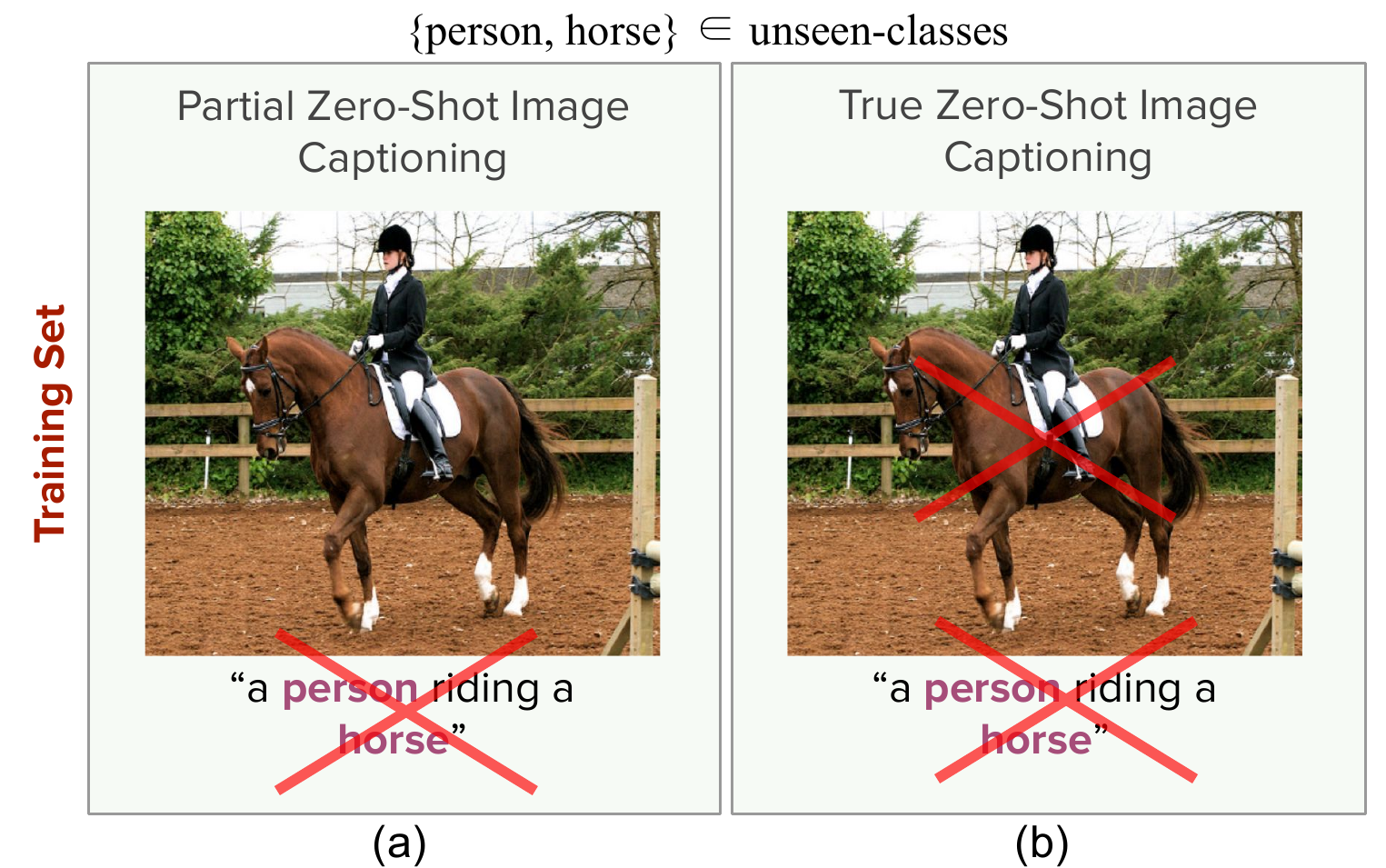}
\end{center}
    \caption{(a) {\em Partial zero-shot image captioning} problem, where the visual examples, without captions, of {\em unseen} classes are used during training. (b) {\em True zero-shot image captioning} problem, neither visual nor textual examples of unseen classes are available during training.}
\label{init}
\end{figure}

In the context of image classification, {\em zero-shot learning} (ZSL) has emerged as a promising alternative towards
overcoming the practical limits in collecting labeled image datasets and constructing image
classifiers with very large object vocabularies. In a similar manner, {\em zero-shot image captioning} (ZSC), aims to develop methods towards
overcoming the data collection bottleneck in image captioning. However, we observe that there is no
prior work irectly tailored
to study captioning in a truly zero-shot setting,
except the preliminary conference version of this paper to the best of our knowledge:
recent works on
ZSC~\cite{lu2018neural, yao2017incorporating} study the ZSC problem only
in the language domain, presuming the availability of a pre-trained fully-supervised object
detector covering all object classes of interest. We refer to these methods as
\textit{partial zero-shot image captioning}.

Following these observations, we propose the problem of {\em true zero-shot captioning}, where test
images contain instances of unseen object categories with no supervised visual or textual examples, in
addition to the seen categories. We believe that this change constitutes a more
direct problem definition towards (i) developing semantically scalable captioning methods, and, (ii)
evaluating captioning approaches in a realistic setting where not all object classes have training
examples. The difference between the partial versus true ZSC problems is
illustrated in Figure~\ref{init}.

\begin{figure*}
\begin{center}
\includegraphics[width=\textwidth]{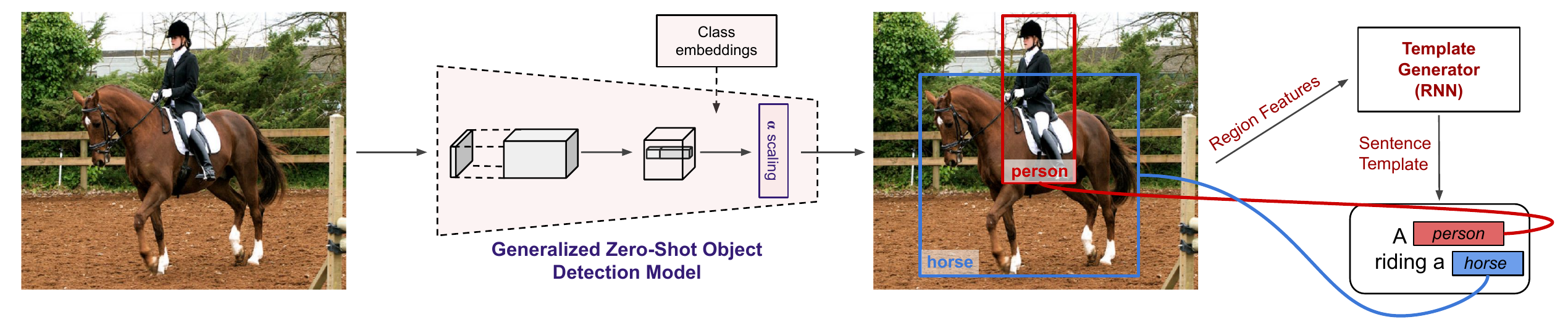}
\end{center}
    \caption{Our zero-shot captioning framework, which consists of two components: (i) a generalized zero-shot object detection model with scaling based score calibration, and (ii) an image caption generation module.}
\label{fig:main}
\end{figure*}

To tackle the true ZSC problem, we propose an approach that consists of a novel
generalized zero-shot detection (GZSD) model, which aims to generate detections in scenes with both
seen and unseen class instances, and a template-based~\cite{lu2018neural} caption generator. A
high-level summary of our ZSC approach can be found in Figure~\ref{fig:main}. 
In order to address the GZSD problem, we propose
a scaling scheme and incorporate {\em uncertainty calibration}~\cite{liu2018generalized} to make
seen and unseen class scores comparable.  We also show out that using class-to-class similarities
obtained over word embeddings~\cite{mikolov2013distributed} 
%between each target class and all training classes 
as {\em class embeddings} improves the GZSD results, compared to using class name
embeddings directly. On the MS-COCO dataset~\cite{lin2014microsoft}, we present a detailed evaluation of
both GZSD and ZSC models. For a more accurate evaluation of the ZSC results, we propose a new
evaluation metric called V(isual)-METEOR, which adapts and improves the widely used METEOR metric
for ZSC evaluation purposes.

A preliminary version of this work has previously appeared in \cite{demirel2019image}. In addition
to provide more detailed related work discussions and method explanations,
this paper extends the conference version by introducing
uncertainty calibration loss for class confidence calibration,
evaluating the impact of various model decisions and score calibration, introducing a comparison
to the recent GZSD methods on the benchmark MS-COCO dataset, 
quantitatively demonstrating the advantage of using class-to-class
similarities as the class embeddings, 
and analyzing the GZSD failure patterns, which are all directly relevant for the captioning quality.
The journal version also proposes the V-METEOR metric, and uses the new metric 
for a more detailed analysis of the ZSC model.

\section{Related Work}
\label{sec:relwork}

Below, we provide an overview of the related work on zero-shot classification, detection and captioning. 

\subsection{Zero-shot classification}

Early work on ZSL focused on directly using attribute based probabilistic models for transferring
knowledge from seen to unseen classes~\cite{lampert09}. More recent works explore other
knowledge transfer mediums and predictive models, \eg \cite{akata2013label,akata2015evaluation, demirel2017attributes2classname, 
changpinyo2016synthesized, kodirov2017semantic, deng2014large,  long2018zero, luo2018zero, yu2018zero}.
A comparative survey of discriminative ZSL models can be found in
Xian~\etal\cite{xian2018zero}, which introduces the problem of {\em generalized zero-shot
learning} (GZSL) problem in an image classification context.
Alternatively, the development of generative models that can synthesize training examples of unseen classes 
has received significant interest in recent years, \eg
\cite{bucher2017generating,felix2018multi,mishra2018generative,xian2019f,zhu2018generative,li2019leveraging,sariyildiz2019gradient,chen2021free,chen2021semantics}.

One of the challenges in GZSL is keeping the seen and unseen class scores
comparable.
A prominent idea in addressing this problem is 
reducing the prediction bias towards seen classes.  For this purpose, Liu
\etal~\cite{liu_generalized_2018} proposes to increase unseen class prediction confidence by minimizing
the entropy of unseen class scores during training. Jian
\etal~\cite{jiang_transferable_2019} promotes higher confidence scores for the {\em familiar} unseen classes during training based on unseen-to-seen class similarity estimates.
Chao \etal~\cite{chao2016empirical} uses an empirically chosen seen class score scaling coefficient. We utilize a similar strategy for GZSD, except that instead of
manually choosing the scaling coefficient, we learn it during training.

\subsection{Zero-shot object detection}

ZSD is a relatively new problem, pioneered by~\cite{rahman2018zero, demirel2018zero, bansal2018zero,
rahman2018polarity, rahman2019transductive, li2019zero, shao2019zero, rahman2020improved, gupta2020multi, li2021inference, zheng2020background, zheng2021visual, nie2022node, yan2022semantics}. These approaches typically extend
supervised detection models to ZSD. Among these studies, Bansal \etal~\cite{bansal2018zero} proposes
a two-step approach that first locates object proposals from low-level
features~\cite{zitnick2014edge} and then classifies the resulting candidate regions using a ZSL
model.
Rahman \etal~\cite{rahman2018zero} proposes a region proposal-based approach and uses a semantic
clustering-based loss term to bring similar classes closer to each other. Demirel
\etal~\cite{demirel2018zero} proposes a regression-based ZSD model that jointly incorporates 
convex combinations of semantic embeddings~\cite{norouzi2014zero} and bi-linear
compatibility models~\cite{akata2013label}.
Rahman \etal~\cite{rahman2018polarity} proposes a polarity loss term that is based on the focal loss approach, to tackle better alignment between visual and semantic domains. Hence, the semantic representations of visually similar classes get closer to each other. 
Li \etal~\cite{li2019zero} uses natural language descriptions of classes for ZSD.
Shao \etal~\cite{shao2019zero} focuses on the candidate proposal generation problem of unseen classes in the ZSD. Gupta \etal~\cite{gupta2020multi} learns a joint embedding space to obtain more discriminative visual and textual embeddings. 
Li \etal~\cite{li2021inference} uses a dual-path method to fuse side analogy information and knowledge transfer between the visual and textual sides. Yan \etal~\cite{yan2022semantics} uses semantics-guided network to improve conventional embeddings.

The model closest to the ZSD component of our ZSC approach is the one proposed by
Demirel \etal~\cite{demirel2018zero}.  Our approach differs by (i) leveraging class-to-class
similarities measured in the word embedding space as class embeddings, as opposed to directly using
the word embeddings, (ii) learning a class score scaling coefficient that reduces the seen class bias and improves GZSD
accuracy, and (iii) exploring the use of uncertainty calibration~\cite{liu2018generalized}
in GZSD. 

There exist alternative learning paradigms that also aim to
reduce the dependency on fully-supervised training examples for object detection.  To this end,
methods for transforming image classifiers into object detectors, \eg~\cite{hoffman2014lsda,
hoffman2015detector,redmon2016yolo9000}, and image-level label based weakly supervised learning
approaches, \eg~\cite{cinbis2016weakly,arun_dissimilarity_2019,ren2020instance}, stand out as closely related directions.
However, such approaches still require labeled training images for all classes of interest, which
can be a major obstacle in building models with the semantic richness needed for captioning.

\subsection{Image captioning} 

State-of-the-art captioning approaches are based on deep neural networks~\cite{mao2014deep,
you2016image, xu2015show, kiros2014unifying, karpathy2015deep, kulkarni2013babytalk, lu2018neural}.
Mainstream methods can be categorized as (i) template-based techniques~\cite{kulkarni2013babytalk,
farhadi2010every, lu2018neural} and (ii) retrieval-based ones~\cite{hodosh2013framing,
ordonez2011im2text,sun2015automatic,kiros2014unifying}. Template-based approaches
generate templates with empty slots, and fill those slots using attributes or detected objects.
Kulkarni \etal~\cite{kulkarni2013babytalk} builds conditional random field models to push tight
connections between the image content and sentence generation process before filling the empty
slots. Farhadi \etal~\cite{farhadi2010every} uses triplets of scene elements for filling the empty
slots in generated templates. Lu \etal~\cite{lu2018neural} uses a recurrent neural network to generate
sentence templates for slot filling. Retrieval-based image captioning methods, in contrast, rely on retrieving
captions from the set of training examples. More specifically, a set of training images similar to
the test example are retrieved and the captioning is performed over their captions.

Dense captioning~\cite{johnson2016densecap,yang2017dense,krishna2017dense} appears to be similar to
ZSC, but the focus is significantly different: while dense captioning
aims to generate rich descriptions, our goal in ZSC  is to achieve captioning over the novel object
classes. Some captioning methods go beyond training with fully supervised captioning data and allow
learning with a captioning dataset that covers only some of the object classes plus additional
supervised examples for training object detectors and/or classifiers for all classes of
interest~\cite{anne2016deep, venugopalan2017captioning,yao2017incorporating, anderson2016guided,
wu2018decoupled}. Since these methods presume that all necessary visual information can be obtained
from some pre-trained object recognition models, we believe they cannot be seen as true ZSC
approaches.

\rev{Recently, the generation of {\em fine-grained captions} has attracted interest~\cite{chen2020say, khan2022deep,
yuan20193g, cheng2020stack}.  Chen \etal~\cite{chen2020say} proposes to use scene graphs to control
caption detail level according to user intentions. 
Khan
\etal~\cite{khan2022deep} uses 
{\em Bahdanau attention}~\cite{bahdanau2014neural} to enrich visual embeddings. 
Yuan \etal~\cite{yuan20193g} proposes gating mechanisms to 
weight global and local cues.  
Cheng \etal~\cite{cheng2020stack} adjusts attention weights of visual feature vectors and semantic feature embeddings in a decoder
cell sequence to obtain rich fine-grained image captions.  Unlike ZSC, these methods do not target
generating captions with objects unavailable in the training set.}

\rev{We additionally study the problem of evaluating ZSC results. 
While various metrics such as METEOR~\cite{denkowski2014meteor} and SPICE~\cite{anderson2016spice} are widely used,
the captioning evaluation is still an open problem, \eg \cite{wang2020towards}.
We propose a ZSC-focused metric that evaluates
the visual and lingual caption quality separately for the unseen and seen classes.}

\section{Method}
\label{sec:method}

In this section, we first explain our 
main ZSD model component, and its GZSD extensions.  We then explain how we build the ZSC model. Finally, we discuss the evaluation difficulties and define the V-METEOR metric.

\subsection{Main zero-shot detection model} In ZSD, the goal is to learn a
detection model over the examples given for the seen classes ($Y_s$) such that the detector can
recognize and localize the bounding boxes of the unseen classes $Y_u$.
For this purpose, we adapt the YOLO~\cite{redmon2016you} architecture to the 
ZSD problem. 

\begin{figure}
\begin{center}
\includegraphics[width=0.75\textwidth]{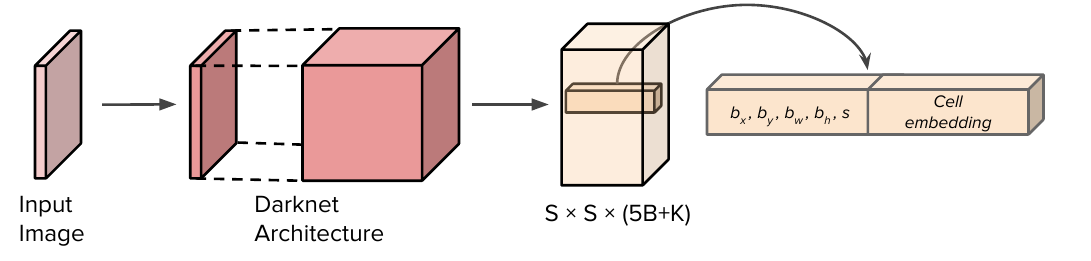}
\end{center}
    \caption{Summary of the proposed GZSD method. At each cell, the network is trained to produce box coordinate predictions (denoted by $b_x,b_y,b_h,b_w$ in the figure), objectness scores (denoted by $s$ in the figure) 
    and a cell embedding to be used for zero-shot recognition.}
\label{zsd-detail}
\end{figure}

In the
original YOLO approach, the loss function consists of three components: (i) the localization loss, which measures the error
between ground truth locations and predicted bounding boxes, (ii) the objectness loss, and (iii) the recognition
loss, over a prediction grid of size $S \times S$.
Following our prior work in \cite{demirel2018zero}, we adapt the YOLO model to the ZSD problem by
replacing per-cell class probability predictions with {\em cell embeddings} and re-defining  the
prediction function as a compatibility estimator between the cell and class embeddings:
\begin{equation} f(x,c,i) = \frac{\Omega(x,i)^{T}\Psi (c)}{\parallel \Omega(x,i) \parallel
\parallel \Psi(c) \parallel } . \label{eq:label_f} \end{equation}
Here, $f(x,c,i)$ is the prediction score corresponding to the class $c$ and
cell $i$, for image $x$, $\Psi(c)$ represents the $c$-th class embedding, and $\Omega(x,i)$ denotes the predicted cell embedding as shown in Figure~\ref{zsd-detail}.
The resulting model, therefore, allows making
detection predictions for samples of novel classes purely based on their class embeddings.

\mypar{Class embeddings} In principle, one can use attributes or word embeddings of class names directly as class embeddings, \eg \cite{demirel2018zero}.  Attributes can provide powerful visual descriptions of classes,
however, they tend to be domain-specific and typically difficult to define for a large variety of
object classes, as needed in ZSC. Word embeddings of class names
are much easier
to collect,  however, they typically contain indirect information about the visual
characteristics of classes, and therefore, known to provide significantly weaker prior knowledge for visual 
recognition~\cite{akata2015evaluation}. 

To use the word embeddings more effectively, we propose to define class embeddings in terms of class-to-class similarities computed over word embeddings: 
we define the class $c$ embedding in terms of the similarity with each seen class $\bar{c}$:
\begin{equation}
    \Psi(c) = \left[ \varphi(c)^{T}\varphi(\bar{c}) + 1 \right]_{\bar{c} \in Y_s}
\label{eq:label_sim}
\end{equation}
where $\varphi(c)$ denotes the $c$-th class name's word embedding. Since semantic relations across classes tend to 
correlate with their visual characteristics, this embedding can provide a valuable implicit visual description 
defined through a series of inter-class similarities. The ZSL method, therefore, can make predictions based collectively on  
these similarity values. We empirically demonstrate the advantage of this scheme in Section~\ref{sec:experiments}.

\subsection{Generalized zero-shot detection extensions}
\label{sec:gzsdextensions}

There can be a significant bias towards the seen classes as the
GZSD model is trained to predict seen class instances.
We use the following two extensions to reduce this bias.

\begin{figure}
\begin{center}
\includegraphics[width=0.85\textwidth]{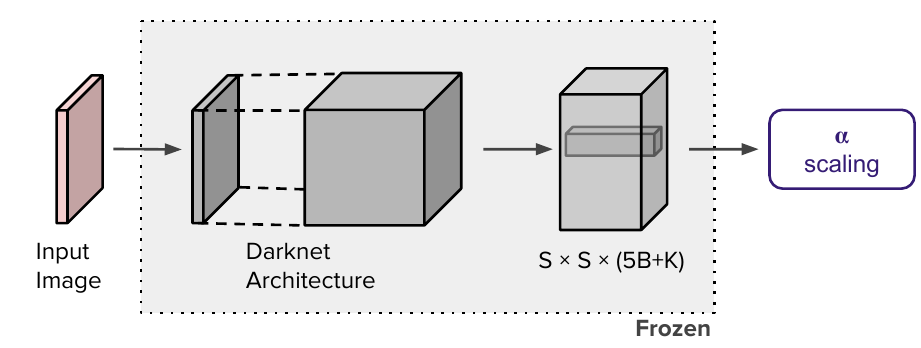}
\end{center}
    \caption{$\alpha$ scaling factor learning process. In the short second round of the learning, all layers of the YOLO-based ZSD method are frozen. The $\alpha$ value is learned through the training set in a few epochs.}
\label{alpha}
\end{figure}

\mypar{Alpha scaling} In this technique, we aim
to 
reduce the bias towards the training classes by
making the unseen and seen class scores more comparable through a score scaling scheme.
For this purpose, we introduce the $\alpha$ coefficient for the unseen
test classes, and redefine $f(x,c,i)$ as follows:
\begin{equation}
f(x,c,i) = \begin{cases}
\alpha \frac{\Omega(x,i)^{T}\Psi (c)}{\parallel \Omega(x,i) \parallel  \parallel \Psi(c) \parallel }, & \text{ if } c \in Y_{u} \\ \\
\frac{\Omega(x,i)^{T}\Psi (c)}{\parallel \Omega(x,i) \parallel  \parallel \Psi(c) \parallel }, & \text{ otherwise }
\end{cases}
\label{eq:label_bias}
\end{equation}
To make the $\alpha$ estimation practical, we want to avoid requiring additional training
examples. For this reason, we first train the ZSD model over all training
classes without $\alpha$. We then designate a subset of seen classes as {\em unseen-imitation}
classes.
To obtain unseen-like confidence scores for these classes, we temporarily set all
entries corresponding to unseen-imitation classes in Eq.~\ref{eq:label_sim} to zeros and treat
unseen-imitation classes as unseen classes in Eq.~\ref{eq:label_bias}.  These modifications allow
us to obtain classification scores as if the model was trained without using the samples of
unseen-imitation classes.  We then train $\alpha$ only, keeping the rest of the
network frozen, as shown in Figure~\ref{alpha}.

Overall, the proposed $\alpha$ coefficient estimation scheme leverages the special structure of our
class embeddings to efficiently approximate the unseen class scores.  While the approximation can
possibly be coarse, we experimentally show in Section~\ref{sec:experiments} that the proposed scheme
is effective for learning the $\alpha$ coefficient, at a negligible extra training cost.

\mypar{Uncertainty calibration} The second unbiasing technique that we explore 
is {\em uncertainty calibration}, adapted from the zero-shot classification approach of Liu
\etal~\cite{liu2018generalized}. The idea is to minimize the uncertainty over
unseen class predictions during training, based on 
the observation that
a prediction model learned over seen class samples tends to yield lower confidence 
scores for unseen classes, resulting in misdetections. 

The uncertainty in confidence scores is quantified via entropy over unseen class probabilities.
We adapt the uncertainty calibration loss $\ell_h$ to our ZSD model as a loss over per-cell predictions:
\begin{equation}
    \ell_{h}(x) = - \sum_{i=0}^{S^{2}} \mathds{1}_\text{obj}^{i} \sum_{c \in Y_{u}} p_u(c|x,i) \log p_u(c|x,i)
\label{eq:uncertainty}
\end{equation}
Here, $p_u(\cdot)$ corresponds to $f(x,c,i)$-driven unseen class likelihoods:
\begin{equation}
p_u(c|x,i) = \frac{\exp(f(x,c,i)/\tau)}{\sum_{c^{'} \in Y_{u}}\exp(f(x,c^{'},i)/\tau)}
\label{eq:uncertainty_dist}
\end{equation}
where $\tau$ denotes the softmax temperature coefficient. $\tau$ is empirically determined as in Liu \etal~\cite{liu2018generalized}. 
The loss encourages more confident unseen class score estimates, as less ambiguous prediction results in smaller entropy values. In order to adapt the uncertainty calibration to the detection model, we first train the ZSD model over all training classes as in the alpha scaling optimization process. We also use the same designated unseen-imitation subset as unseen classes. In the second training stage, we temporarily set all entries corresponding to unseen-imitation classes to zeros and then fine-tune the whole model without freezing any layers, unlike alpha scaling coefficient learning. 

\subsection{Zero-shot captioning model}
\label{sec:zsc:approach}

Our goal is the construction of an image captioning model that can accurately summarize scenes
potentially with seen and unseen class instances.  For this purpose, we opt to use a template-based
captioning method which provides the sentence templates with visual word slots to be filled
based on the outputs of an object detection model. 

We adapt the slotted sentence template generation model of
{\em Neural Baby Talk} (NBT)~\cite{lu2018neural}. The NBT method generates sentence templates
which consist of the empty word slots by using a recurrent neural network.  To obtain a
content-based attention mechanism over the grounding regions, NBT embraces {\em pointer
networks}~\cite{vinyals2015pointer}.
The NBT model is trained by optimizing the model
parameters $\omega$ such that  the log-likelihood of each ground-truth caption $q$ conditioned on the corresponding image $x$ is maximized:
\begin{align}
\omega^* = \text{arg}\,\max\limits_{\omega}\,\sum_{(x,q)}^{}\log p(q | x; \omega) .
\end{align}
Here, the conditional caption likelihood $p(q | x; \omega)$ of $|q|$ words is measured auto-regressively, using a recurrent network:
\begin{align}
p(q | x;\omega) = \prod_{t=1}^{|q|}p(q_t | q_{1:t-1}, x; \omega) .
\end{align}
The NBT method additionally incorporates a latent variable $r_t$ to represent the specific image region, so
the probability of a word $q_t$ is modeled as follows:
\begin{align}
p(q_t | q_{1:t-1}, x; \omega) = p(q_t | r_t, q_{1:t-1}, x; \omega) p(r_t | q_{1:t-1}, x; \omega) .
\end{align}
The NBT defines two word types for $q_t$, corresponding to {\em textual} and {\em visual} words.
Textual words are not directly related to any image region or specific visual object instance,
therefore the model provides only dummy grounding for them. The template generation network uses the
object detection outputs to fill empty visual word slots, where we utilize the outputs of our GZSD
model.

We train both the GZSD model and the sentence template generation component of NBT
over examples containing only the seen class instance annotations, as required by the {\em true
ZSC} protocol. At test time, we use the GZSD outputs over all classes 
as inputs to the NBT sentence generator. 

\subsection{Measuring zero-shot captioning quality}
\label{sec:zsc:metrics}

{\em Partial zero-shot image captioning} approaches use existing captioning metrics, such as
METEOR~\cite{denkowski2014meteor}, SPICE~\cite{anderson2016spice} and F1 score, for evaluation
purposes.  While these generic textual similarity based metrics provide useful information about the quality of captioning results,
they do not explicitly handle the problem of capturing
visual content within the generated sentence. Therefore, such metrics can possibly be heavily influenced
by structural and syntactic similarities across generated and ground-truth sentences.
Exceptionally, F1 score differs in this regard by completely ignoring the sentence structure and
measuring only the coverage of (unseen) class names within captions. However, F1 score fails
to measure the overall quality or accuracy of the generated sentences, which is also clearly important.

We observe that, based on our experiments in Section~\ref{sec:experiments}, the explicit
handling of visual and non-visual content in the evaluation of sentences is particularly necessary
for {\em true zero-shot image captioning}. In this setting, the problem of generating sentences
that summarize the visual content accurately, including visual entities that are completely unseen during
training, is fundamentally challenging, especially in comparison to partial ZSC with
fully-supervised visual recognition models. Therefore, we propose a new captioning
evaluation metric as a step towards formalizing better metrics for true ZSC.

We develop our metric based on METEOR, which is known to be a simple yet effective metric that
yields a strong correlation with human judgment~\cite{kilickaya2016re}. The original METEOR metric is defined 
by the following formula:
\begin{equation}
    \text{METEOR}=F_\text{mean}(1-p)
\label{eq:meteor}
\end{equation}
where $F_\text{mean}$ aims to capture correctness in terms of unigram precision and recall values and $p$ 
is a penalty term for evaluating the overall sentence compatibility.
More specifically, $F_\text{mean}$ is given by:
\begin{equation}
F_\text{mean}=\frac{10PR}{R + 9P}
    \label{eq:fmean}
\end{equation}
where $P$ and $R$ are the unigram precision and unigram recall values, respectively. These are calculated as:
\begin{eqnarray}
    P &=& \frac{m}{w_{t}} \\
    R &=& \frac{m}{w_{r}}
\end{eqnarray}
where $m$ is the number of unigrams in both reference and generated captions, $w_{t}$ is the number of unigrams in the candidate caption
and $w_{r}$ is the number of unigrams in the reference caption.  The $p$ penalty term checks
how well textual chunks match between a pair of reference and generated captions, using the following definition:
\begin{equation}
p = 0.5 \left ( \frac{c}{u_{m}} \right ) ^{3}
\end{equation}
where $c$ is number of maximally long matching subsequences, and $u_{m}$ is number of mapped unigrams. 

We extend the METEOR metric by defining two separate $F_\text{mean}$ metrics for the visual and non-visual entities. 
For this purpose,
we compute $F_\text{mean}^{v}$ and $F_\text{mean}^{n}$, similar to Eq.~\ref{eq:fmean}, separately over only visual words and only non-visual words, respectively.
We, then, define the proposed metric V-METEOR based on their harmonic mean, as follows:
\begin{equation}
 \text{V-METEOR}=\frac{2F_\text{mean}^{v}F_\text{mean}^{n}}{F_\text{mean}^{v}+F_\text{mean}^{n}}(1-p)
\label{eq:vismeteor}
\end{equation}
In this manner, the proposed V-METEOR metric explicitly measures the joint visual or non-visual accuracy of a sentence, through the
harmonic mean of the $F_\text{mean}^{v}$ and $F_\text{mean}^{n}$ terms. It also incorporates the overall sentence similarity by keeping the
penalty term ($p$) as in METEOR. 

To be able to measure per-class captioning quality, which is particularly valuable in the ZSC
context, we separately compute V-METEOR for each class. In the calculation of the V-METEOR score
of a sentence for a class, the words corresponding to the class name are considered as the visual
words, and the words that are not corresponding to any one of the class names are considered as
non-visual words.  The overall V-METEOR score is obtained by averaging per-class scores.

Finally, we additionally define the following two variations for separately measuring the visual and non-visual quality of the generated sentences, respectively:
\begin{eqnarray}
    &\text{V-METEOR}\textsubscript{vis} = F_\text{mean}^{v}(1-p) \\
    &\text{V-METEOR}\textsubscript{nvis} = F_\text{mean}^{n}(1-p)
\end{eqnarray}
We use $\text{V-METEOR}\textsubscript{vis}$ and $\text{V-METEOR}\textsubscript{nvis}$ 
to gain additional insights.

\section{Experiments}
\label{sec:experiments}

In this section, we explain our experimental setup, present the GZSD and ZSC results,
discuss the V-METEOR evaluations, and provide additional analyses.

\subsection{Experimental setup}
\label{sec:expsetup}

ZSD and (partial) ZSC works use different splits of the MS-COCO dataset for historical reasons. To make our results comparable to related works, we use the same splits as in the related works, separately for GZSD and ZSC as explained below.

\mypar{GZSD evaluation} We use MS-COCO~\cite{lin2014microsoft} dataset in our experiments. 
In our main GZSD experiments, we use the same dataset splits and settings as in the recent work~\cite{gupta2020multi, li2021inference, zheng2020background, zheng2021visual, nie2022node, yan2022semantics, rahman2020improved, rahman2019transductive, demirel2018zero}, 
where $15$ of $80$ MS-COCO classes are used as unseen classes. There also exist different ZSD methods (\eg SB~\cite{bansal2018zero} and DSES~\cite{bansal2018zero}), but they use only 48/17 seen-unseen class distribution or do not share GZSD results with 65/15, so we do not report any comparisons with these methods.

\mypar{ZSC evaluation} For the ZSC approach, we compare the proposed approach with selected {\em upper-bound} methods from \cite{anne2016deep, venugopalan2017captioning,yao2017incorporating, anderson2016guided, lu2018neural}. We again use the same dataset splits and settings as in these works,
where $8$ of $80$ MS-COCO classes are used as the unseen classes.
 
\mypar{Word embeddings} For the GZSD model, we use 300-dimensional
word2vec~\cite{mikolov2013efficient} class name embeddings. For the names containing more
than one word, \eg~{\em tennis racket}, we take the average of the per-word embeddings.  We use
300-dimensional GloVe vector embeddings~\cite{pennington2014glove} in the template generation
component of the ZSC, following the NBT approach~\cite{lu2018neural}.

\subsection{Generalized zero-shot object detection}
\label{sec:detection}

In this section, we report and discuss experimental results for the GZSD model. We train the model for 160 epochs with a
learning rate of $0.001$, and a batch size of $32$. 
Once the model is trained, we
select $8$ out of $65$ seen classes as {\em unseen-imitation} classes for alpha scaling optimization
and uncertainty calibration purposes,
and continue training for $10$ more epochs.

\mypar{Main results} We present the experimental results in Table~\ref{table:gzsd_results}. The upper part of the
table presents results of the two-stage object detection techniques,
and the lower part 
presents the single-stage techniques and our approach, which we call {\em SimEmb}. In the lower part,
\textbf{SimEmb-base}, which represents the model without score calibration,
obtains $28.54\%$ mAP on seen classes, $12.45\%$
mAP on unseen classes and $17.34$ harmonic mean (HM). \textbf{SimEmb}, which
represents the version with learned $\alpha$ scaling coefficient, obtains $28.91\%$ mAP on seen classes, $15.78\%$ mAP on unseen classes and
$20.41\%$ HM.  Finally, \textbf{SimEmb*} represents an upper-bound reference model,
where alpha scaling coefficient is empirically tuned on the test set to maximize the HM score by
evaluating for a range of $\alpha$ values. This upper-bound model
obtains $28.87\%$ mAP on seen classes, $16.00\%$ mAP on unseen classes, and $20.59$ HM value. 

From the results, we first observe that our single-stage approach improves the state-of-the-art among single-stage GZSD models.
We also observe that SimEmb performs similar to or better than many two-stage GZSD models, with the only exception being the very recently published
two-stage approach ContrastZSD~\cite{yan2022semantics}.
Second, the improvements obtained by SimEmb show that
alpha scaling coefficient is crucial for obtaining higher accuracy on unseen class detections
and alpha scaling does not disrupt the seen class performance.  Finally, the comparison between SimEmb and
the SimEmb* upper-bound shows that the proposed alpha scaling learning scheme is effective as
it yields results comparable to directly tuning $\alpha$ on the test set.

\begin{table}[]
\centering
\begin{tabular}{cc|ccc}
\toprule
Category & Method                    & seen           & unseen         & HM             \\ \midrule
\multirow{8}{*}{two-stage}    & MS-Zero~\cite{gupta2020multi}        & \textbf{42.40} & 12.90          & 19.79          \\
                              & MS-Zero++~\cite{gupta2020multi}      & 35.00          & 13.80          & 19.78          \\
                              & DPIF-S~\cite{li2021inference}         & 32.72          & 13.95          & 19.56          \\
                              & DPIF-M~\cite{li2021inference}         & 29.33          & 16.36          & 21.00          \\
                              & BLC~\cite{zheng2020background}            & 36.00          & 13.10          & 19.20          \\
                              & VL-SZSD~\cite{zheng2021visual}        & 39.45          & 13.18          & 19.76          \\
                              & FNG~\cite{nie2022node}             & 38.10          & 13.90          & 20.40          \\
                              & ContrastZSD~\cite{yan2022semantics}    & 40.20          & \textbf{16.50} & \textbf{23.40} \\ \midrule
\multirow{6}{*}{single-stage} & TL~\cite{rahman2019transductive}             & 28.79          & 14.05          & 18.89          \\
                              & PL~\cite{rahman2020improved}              & \textbf{34.07} & 12.40          & 18.18          \\
                              & HRE~\cite{demirel2018zero}            & 28.40          & 12.80          & 17.65          \\
                              & SimEmb-base & 28.54          & 12.45          & 17.34          \\
                              & SimEmb         & 28.91          & 15.78          & 20.41          \\
                              & SimEmb*        & 28.87          & \textbf{16.00} & \textbf{20.59}
                              \\
                              \bottomrule
\end{tabular}
    \caption{mAP results on MS-COCO dataset with GZSD (65/15) settings. \textbf{SimEmb-base}, \textbf{SimEmb}
and \textbf{SimEmb}* correspond to our model without confidence calibration, with learned $\alpha$, and with optimal
$\alpha$ (upper-bound), respectively. \label{table:gzsd_results}}
    
\end{table}

We also observe that the proposed model achieves results comparable to those of two-stage approaches. While
single-stage and two-stage detectors are built on very different design principles and trade-offs,
the overall competitiveness is noteworthy since the work on other low-shot detection problems show that
two-stage models typically yield higher AP scores~\cite{li2021beyond}.

\begin{figure*}
\centering
\begin{tabular}{ccccccccc}
\HangBox{\includegraphics[width=2.55cm, height=2.55cm]{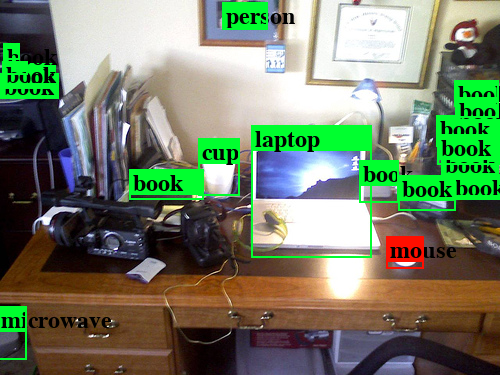}}
\HangBox{\includegraphics[width=2.55cm, height=2.55cm]{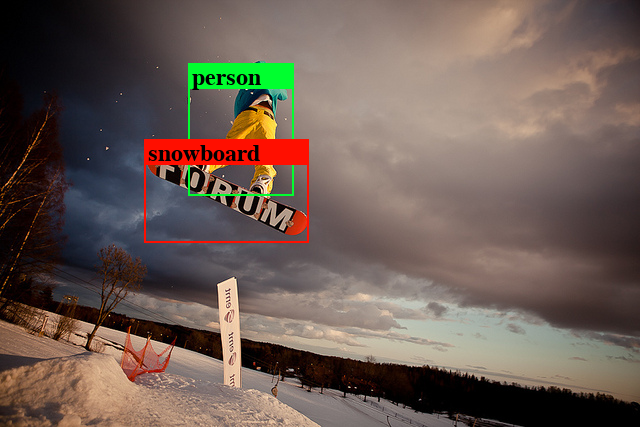}}
\HangBox{\includegraphics[width=2.55cm, height=2.55cm]{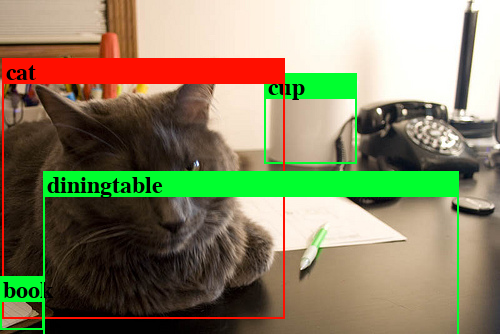}}
\HangBox{\includegraphics[width=2.55cm, height=2.55cm]{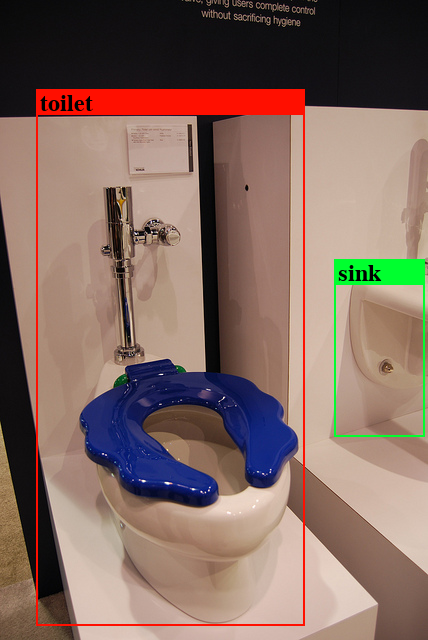}}
\HangBox{\includegraphics[width=2.55cm, height=2.55cm]{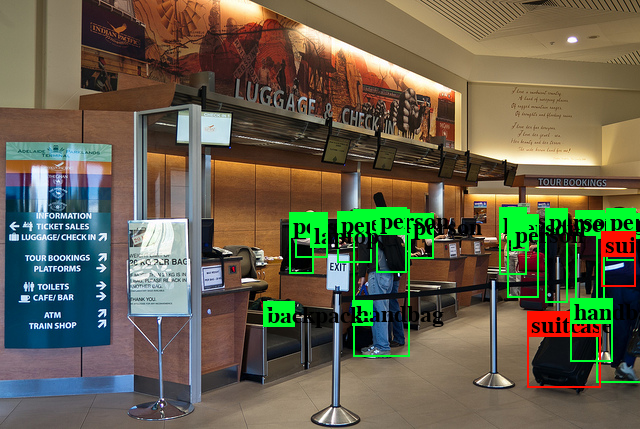}}
\\
\HangBox{\includegraphics[width=2.55cm, height=2.55cm]{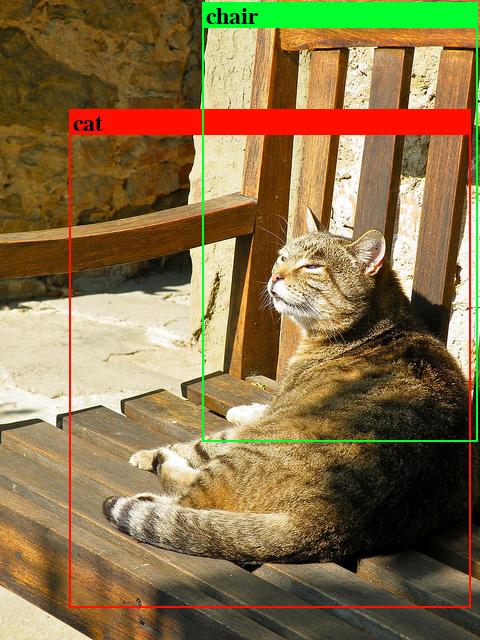}}
\HangBox{\includegraphics[width=2.55cm, height=2.55cm]{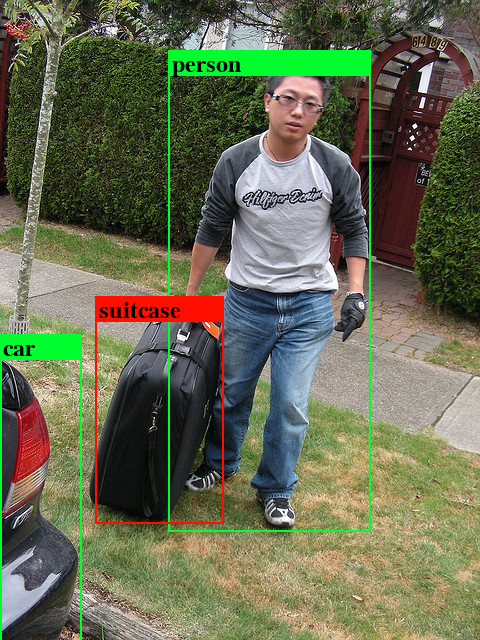}}
\HangBox{\includegraphics[width=2.55cm, height=2.55cm]{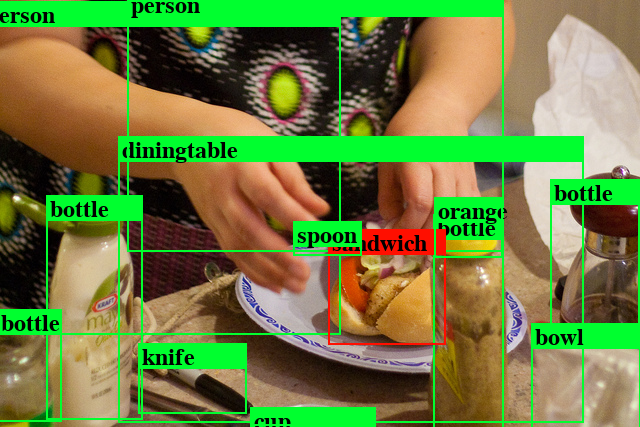}}
\HangBox{\includegraphics[width=2.55cm, height=2.55cm]{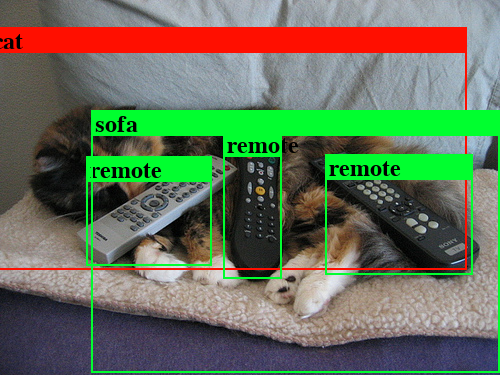}}
\HangBox{\includegraphics[width=2.55cm, height=2.55cm]{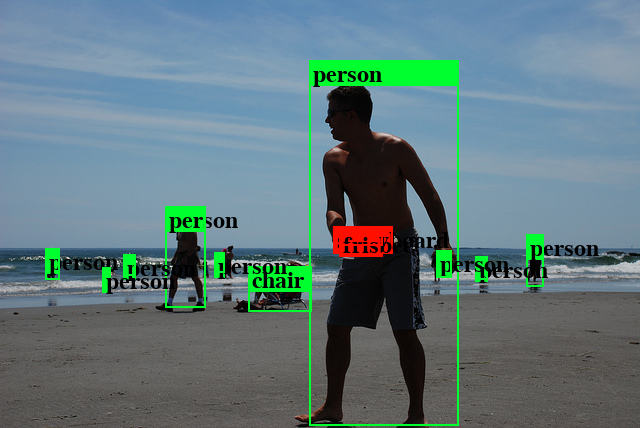}}
\end{tabular}
\caption{GZSD results on scenes containing various \textbf{\textcolor{green}{seen}} and \textbf{\textcolor{red}{unseen}} class instances. (Best viewed in color.)}
\label{fig:gzsd-visual}
\end{figure*}

Qualitative detection results using the proposed SimEmb model can be found in
Figure~\ref{fig:gzsd-visual}. 

\begin{figure}
\begin{center}
\includegraphics[width=0.6\textwidth]{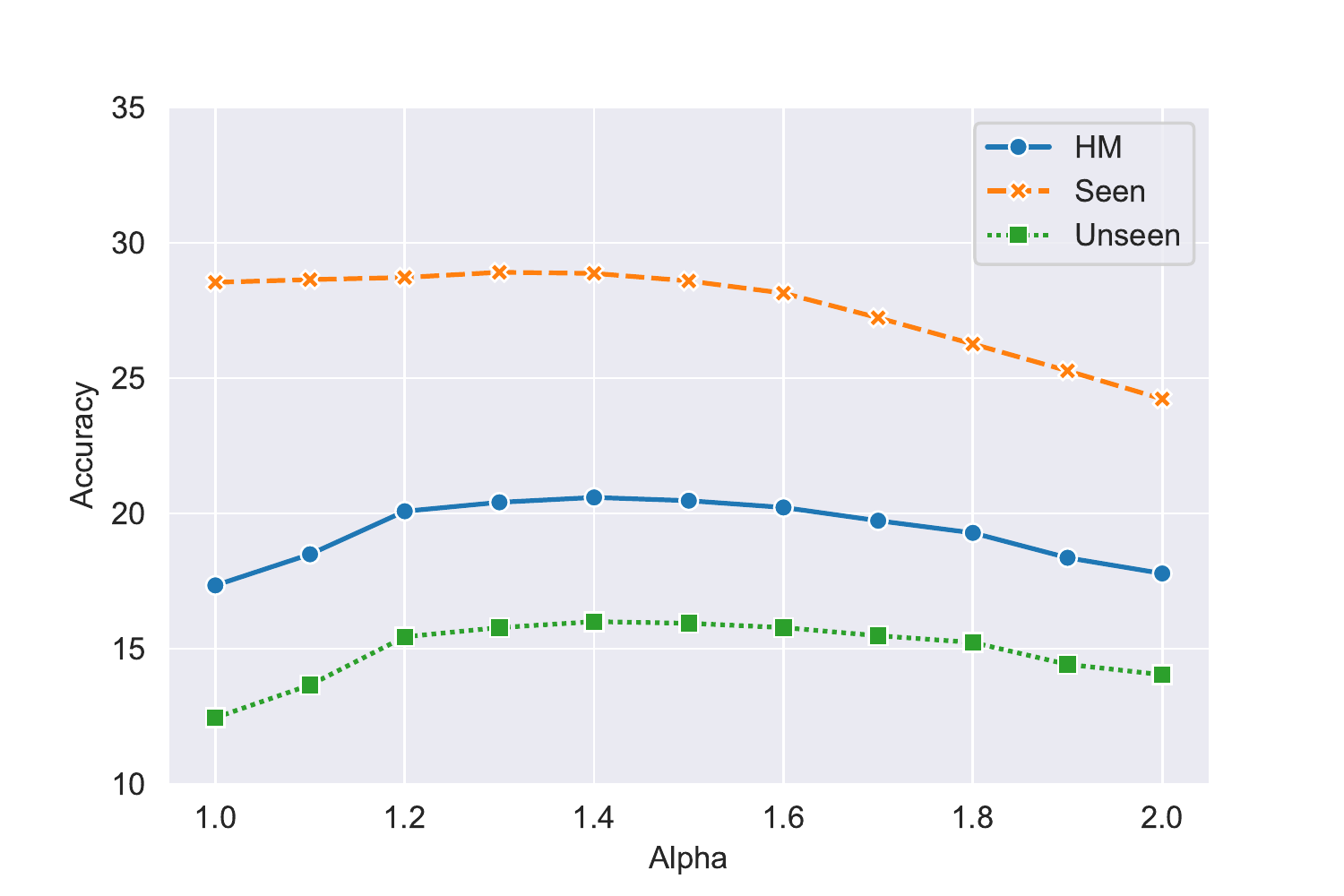}
\end{center}
    \caption{The accuracy values of the proposed method in the GZSD test splits of MS-COCO according to different alpha scaling factors.\label{fig:alpha}}
\end{figure}

\mypar{Correctness of $\alpha$ estimation} We present the evaluation results as a function of $\alpha$ in
Figure~\ref{fig:alpha}. We observe that the best empirical $\alpha$ coefficient
value (in HM) among the tested ones is $1.4$. The proposed $\alpha$ estimator, which
in contrast uses only training examples, results in $\alpha=1.28$, which is both value-wise and
performance score-wise close to the optimal choice.

\mypar{Alpha scaling versus uncertainty calibration} 
As an alternative to alpha scaling for GZSD, we evaluate the uncertainty calibration technique, as explained in 
Section~\ref{sec:gzsdextensions}. 
\rev{We} present the results in Table~\ref{table:calibration_results}, with the following combinations from top to the bottom: base model,
uncertainty calibration ({\em uc-calib}) only, 
alpha scaling only, and their combination. We observe that uncertainty calibration alone performs poorly probably due to 
the difficulty of correcting class bias purely based on fine-tuning. Our alpha scaling technique yields a much better
result in terms of HM score, with an improvement from $17.34$ to $20.41$. The combination of the two techniques slightly improves
the HM score to $20.46$. This proves that the alpha scaling scheme is effective in comparison to a state-of-the-art 
calibration technique. 
For the sake of simplicity, we keep using only alpha scaling in our following experiments. 

\begin{table}[]
\centering
\begin{tabular}{lcccc}
\toprule
$\alpha$-scaling           & uc-calib        & seen           & unseen         & HM                    \\
\midrule
                           &                                 & 28.54          & 12.45          & 17.34\\
                           & \checkmark                      & 28.60          & 11.15          & 16.04 \\
\checkmark                 &                                 & \textbf{28.91} & 15.78          & 20.41\\
\checkmark                 & \checkmark                      & 28.85          & \textbf{15.85} & \textbf{20.46} \\
\bottomrule
\end{tabular}
    \caption{mAP results on MS-COCO dataset in the 65/15 GZSD setting, using the base model with and without alpha scaling and uncertainty calibration ({\em uc-calib}).}
\label{table:calibration_results}
\end{table}

\begin{table*}
\centering

\resizebox{\textwidth}{!}{%
\begin{tabular}{l|l|cccccccc|ccc}
\toprule
Exp. Type        & Test & \multicolumn{1}{l}{bottle} & \multicolumn{1}{l}{bus} & \multicolumn{1}{l}{couch} & \multicolumn{1}{l}{microwave} & \multicolumn{1}{l}{pizza} & \multicolumn{1}{l}{racket} & \multicolumn{1}{l}{suitcase} & \multicolumn{1}{l|}{zebra} & \multicolumn{1}{l}{U-mAP(\%)} & \multicolumn{1}{l}{S-mAP(\%)} & \multicolumn{1}{l}{HM} \\ 
\midrule
ZSD           & U    & 5.2                        & 53.3                    & 35.1                      & 23.9                          & 44.4                      & 36.4                       & 9.1                          & 43.7                       & 31.4                          & -                             & -                      \\ 
\midrule
GZSD w/o $\alpha$ & S+U  & 0                          & 0                       & 2.7                       & 0                             & 0                         & 0                          & 0                            & 0                          & 0.3                           & 27.4                          & 0.7                    \\
GZSD           & S+U  & 0.8                        & 21.4                    & 4.9                       & 1.2                           & 4.8                       & 0.7                        & 9.1                          & 15.8                       & 7.3                           & 19.2                          & 10.6      \\
\bottomrule
\end{tabular}%
}
\caption{Our results on ZSD and GZSD (72/8). The first row represents the experimental results where we only use images belonging to the unseen classes and unseen class embeddings, the remaining rows represent the GZSD results where we use all class embeddings on the MS-COCO val5k split.}
\label{table:gzsd_72_8}
\end{table*}

\mypar{GZSD results on ZSC splits} In our experiments presented so far, we have used the 65/15 COCO split.
In our ZSC experiments, however, we need to use the alternative 72/8 split of \cite{anne2016deep} to make comparisons to the related work.
Therefore, here we report the results of our GZSD model on the 72/8 split. We train the model using the same hyper-parameters as before.
We select $8$ out of $72$ seen classes as unseen-imitation classes for alpha scaling optimization.

We evaluate the detection model under the ZSD and GZSD scenarios.  For the ZSD
experiments, we use the MS-COCO validation images consisting of unseen class instances.  For the
GZSD experiments, we use the whole MS-COCO val5k split. We present the results on
Table~\ref{table:gzsd_72_8}.  In the ZSD case, we observe an unseen class mAP of
$31.4\%$.  In the GZSD case, we observe a much lower $0.3\%$ mAP without alpha scaling,
and $0.7$ HM. Alpha scaling improves the unseen class mAP to $7.3\%$ and the HM score to $10.6$.  We
note that prior works on GZSD do not use this ZSC (72/8) split, therefore, we do not report any
comparisons to the state-of-the-art in this split.
We also note that our primary interest in GZSD is to build a strong method to serve as a crucial
component of ZSC, therefore, these results highlight one of the major difficulties
in building accurate captioning models in the realistic ZSC setting.

\subsection{Zero-Shot image captioning}
\label{sec:exp:caption}

For the ZSC experiments, we use 
the same experimental setup described in~\cite{lu2018neural}, and exclude the image-sentence pairs
containing unseen class instances during training. We consider the partial ZSC
approaches proposed in \cite{anne2016deep, venugopalan2017captioning,yao2017incorporating,
anderson2016guided, lu2018neural}
as upper-bound baselines for our true ZSC setting.  We also define and evaluate a
baseline method based on NBT, where we train the NBT captioning model based solely on the training
classes without integrating our GZSD model. We refer to this model as {\em NBT-baseline}. 

To establish a fair comparison, we follow the practices of the NBT~\cite{lu2018neural} approach. We
evaluate the ZSC model on the selected validation subset of the MS-COCO caption dataset.  To obtain
per-class evaluation scores, we use the F1 metric~\cite{anne2016deep}, where a visual class is considered as relevant
in an image if that class name appears in any one of the human generated reference captions for that image, and irrelevant otherwise.
Similarly, on a test image, a model-generated caption is considered as correct for a visual class if the generated caption includes
(excludes) the corresponding word for that relevant (irrelevant) class.
The per-class F1 score is then defined as the ratio of correctly captioned test images.
We additionally use the well-established METEOR~\cite{denkowski2014meteor} and
SPICE~\cite{anderson2016spice} metrics, in addition to averaging the per-class F1 scores (referred to as {\em Avg. F1}).
We separately discuss the evaluation results in terms of the proposed V-METEOR metric in the
next section. 

\begin{table*}[]
\centering

\resizebox{\textwidth}{!}{%
\begin{tabular}{l|cccccccc|ccc}
\toprule
Method     & \multicolumn{1}{l}{bottle} & \multicolumn{1}{l}{bus} & \multicolumn{1}{l}{couch} & \multicolumn{1}{l}{microwave} & \multicolumn{1}{l}{pizza} & \multicolumn{1}{l}{racket} & \multicolumn{1}{l}{suitcase} & \multicolumn{1}{l|}{zebra} & \multicolumn{1}{l}{Avg. F1} & \multicolumn{1}{l}{METEOR} & \multicolumn{1}{l}{SPICE} \\
\midrule
\multicolumn{12}{c}{True zero-shot captioning}\\
\midrule
NBT-baseline  & 0& 0& 0& 0& 0& 0& 0& 0& 0 & 18.2 & 12.7 \\
Our method & 2.4                       & 75.2                    & 26.6                      & 24.6                          & 29.8                      & 3.6                       & 0.6                         & 75.4                       & 29.8                       & 21.9                       & 14.2 \\
\midrule
\multicolumn{12}{c}{Partial zero-shot captioning (upper-bounds)}\\
\midrule
DCC~\cite{anne2016deep}& 4.6                        & 29.8                    & 45.9                      & 28.1                          & 64.6                      & 52.2                       & 13.2                         & 79.9                       & 39.8                       & 21.0                       & 14.4                      \\
NOC~\cite{venugopalan2017captioning}& 17.8                       & 68.8                    & 25.6                      & 24.7                          & 69.3                      & 68.1                       & 39.9                         & 89.0                       & 49.1                       & 21.4                       & -                         \\
C-LSTM~\cite{yao2017incorporating}& 29.7                       & 74.4                    & 38.8                      & 27.8                          & 68.2                      & 70.3                       & 44.8                         & 91.4                       & 55.7                       & 23.0                       & -                         \\
Base+T4~\cite{anderson2016guided}& 16.3                       & 67.8                    & 48.2                      & 29.7                          & 77.2                      & 57.1                       & 49.9                         & 85.7                       & 54.0                       & 23.3                       & 15.9                      \\
NBT+G~\cite{lu2018neural}& 14.0                       & 74.8                    & 42.8                      & 63.7                          & 74.4                      & 19.0                       & 44.5                         & 92.0                       & 53.2                       & 23.9                       & 16.6                      \\
DNOC~\cite{wu2018decoupled}& 33.0                       & 77.0                    & 54.0                      & 46.6                          & 75.8                      & 33.0                       & 59.5                         & 84.6                       & 57.9                       & 21.6                       & -                            \\
\bottomrule
\end{tabular}%
}

\caption{Zero-shot captioning results with comparison to captioning models involving visually fully-supervised models. \label{table:captioning}}
\end{table*}

We present the results in Table~\ref{table:captioning}. First, we observe that the proposed approach
greatly outperforms the NBT-baseline with clear improvements in terms of Avg. F1 ($0$ to $29.8$),
METEOR ($18.2$ to $21.9$) and SPICE ($12.7$ to $14.2$) scores. This shows the value of explicitly
handling the GZSD task as part of the captioning process. In comparison to the upper-bound
partial-ZSC captioning approaches, which involve supervised visual training in both seen and unseen classes, our
approach yields comparable results in terms of METEOR and SPICE metrics. In particular, we observe that the ZSC model yields better
results compared to the DCC~\cite{anne2016deep} and NOC~\cite{venugopalan2017captioning} methods. This is most probably due to the fact that our sentence
template generation method provides accurate locations for visual words, enabling the generation of
more natural and visually grounded captions. We observe 
relatively lower scores for the ZSC model, compared to the remaining supervised models.

\begin{figure*}
\centering
\resizebox{1\textwidth}{!}{%    
\begin{tabular}{ccccccccc}
\HangBox{\includegraphics[width=3.8cm, height=3.4cm]{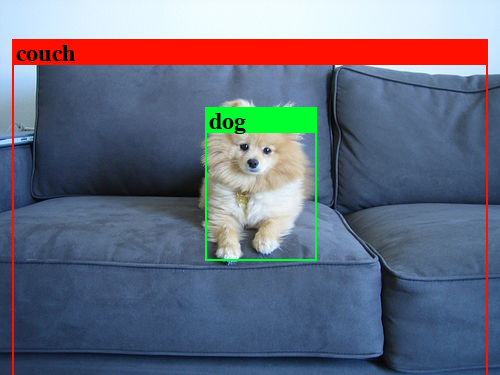}\\
A small white \textit{\textcolor{green}{dog}} sitting\\ on a \textbf{\textcolor{red}{couch}}.}
\HangBox{\includegraphics[width=3.8cm, height=3.4cm]{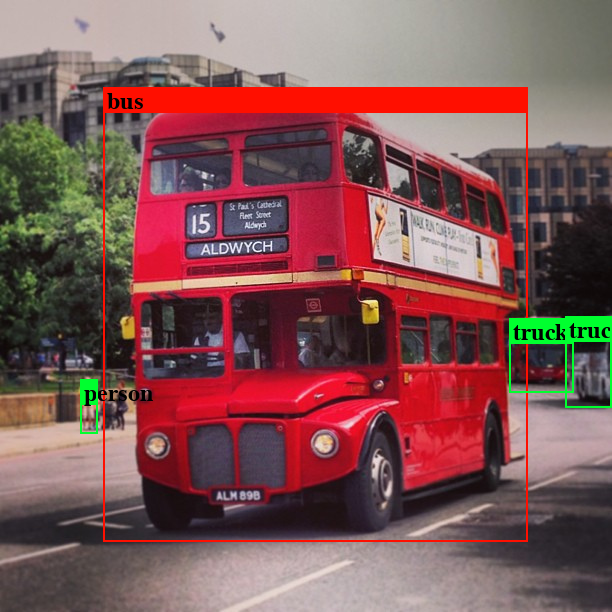}\\
A red \textbf{\textcolor{red}{bus}} is driving\\ down the street.}
\HangBox{\includegraphics[width=3.8cm, height=3.4cm]{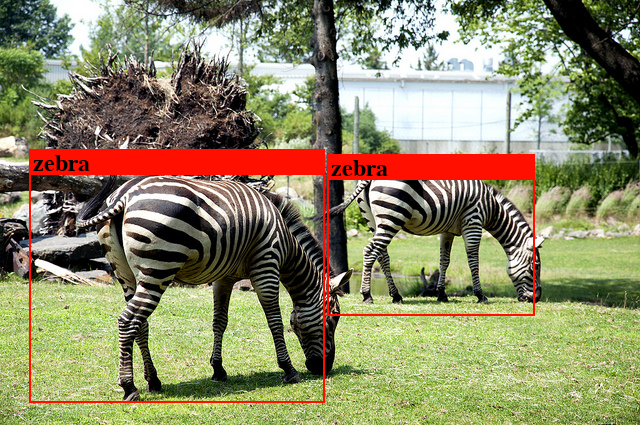}\\
A couple of \textbf{\textcolor{red}{zebra}}\\ standing in a field.}
\HangBox{\includegraphics[width=3.8cm, height=3.4cm]{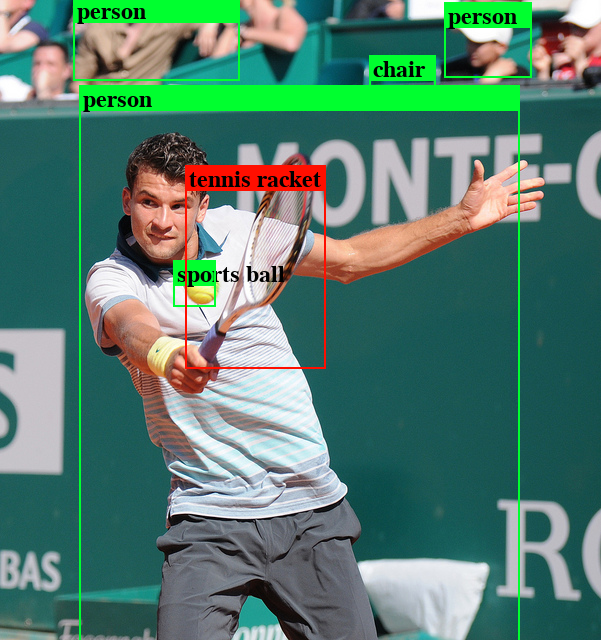}\\
A \textit{\textcolor{green}{tennis player}} is about\\ to hit a \textbf{\textcolor{red}{tennis racket}}.}
\HangBox{\includegraphics[width=3.8cm, height=3.4cm]{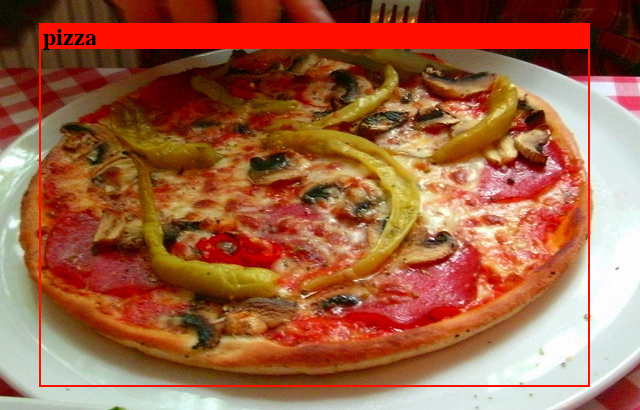}\\
A white plate topped\\ with a piece of \textbf{\textcolor{red}{pizza}}.}
\\
\HangBox{\includegraphics[width=3.8cm, height=3.4cm]{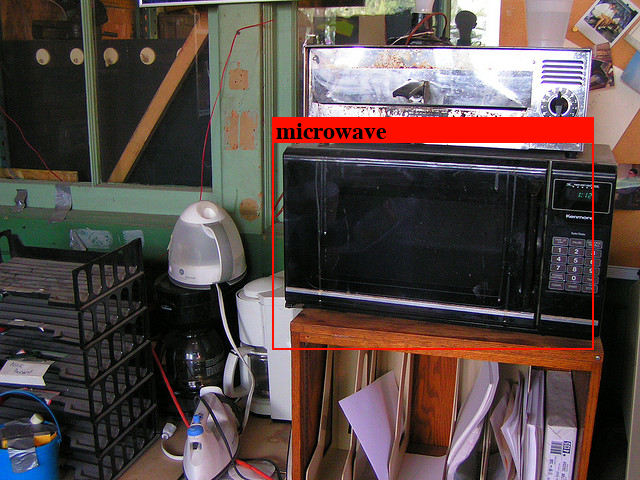}\\
A kitchen with a \textbf{\textcolor{red}{m.wave}}\\ and a counter.}
\HangBox{\includegraphics[width=3.8cm, height=3.4cm]{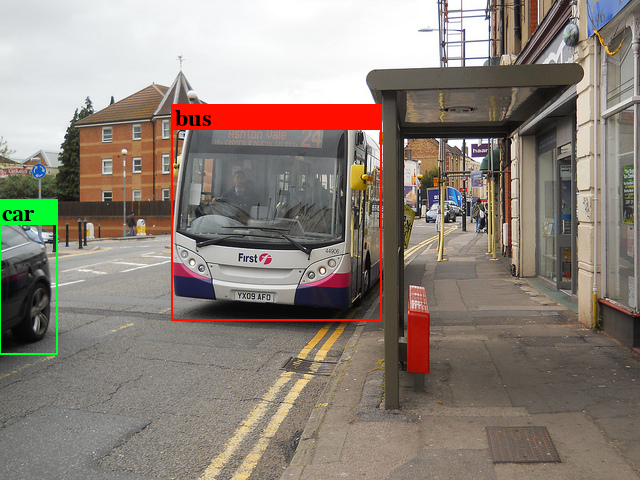}\\
A \textbf{\textcolor{red}{bus}} is parked on the\\ side of the street.}
\HangBox{\includegraphics[width=3.8cm, height=3.4cm]{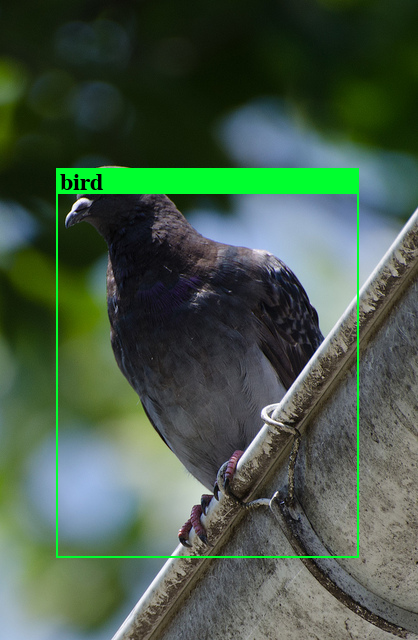}\\
A \textit{\textcolor{green}{bird}} sitting on top of a\\ metal pole.}
\HangBox{\includegraphics[width=3.8cm, height=3.4cm]{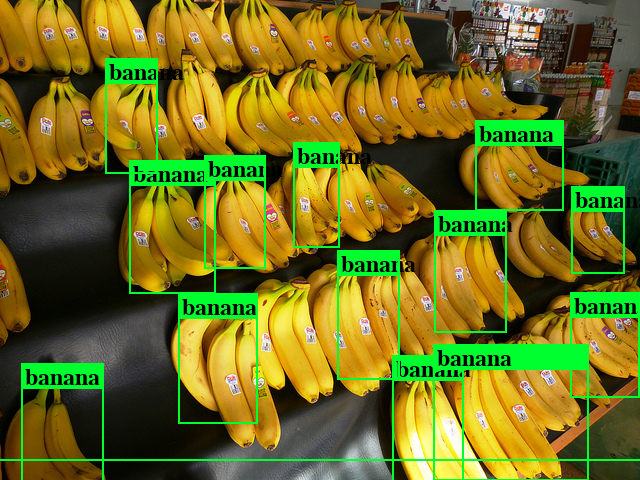}\\
A bunch of \textit{\textcolor{green}{banana}} that\\ are on a table.}
\HangBox{\includegraphics[width=3.8cm, height=3.4cm]{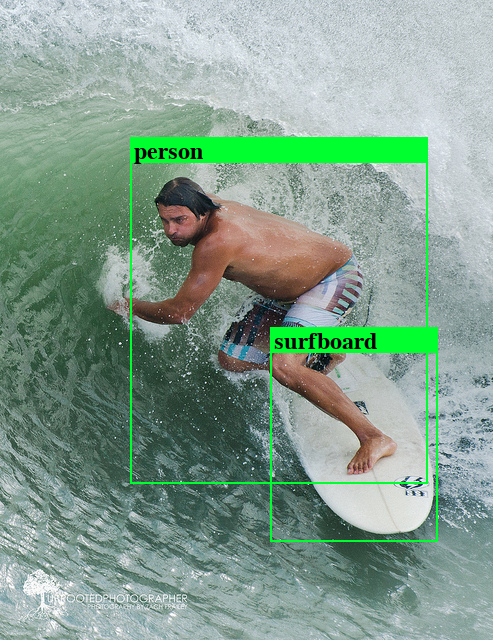}\\
\textit{\textcolor{green}{man}} riding a wave on\\ top of a \textit{\textcolor{green}{surfboard}}.}
\end{tabular}}
\caption{Image captioning results on images with \textit{\textcolor{green}{seen}} and \textbf{\textcolor{red}{unseen}} class instances. (Best viewed in color.)}
\label{fig:zscresults}
\end{figure*}

Noticeably, the performance gap between true ZSC and (visually) supervised partial ZSC is larger
in terms of the Avg. F1 metric.  This is mostly an expected result as the F1 metric directly measures the ability
to incorporate visual classes during captioning, akin to a visual recognition metric.  Here,
supervised methods are known to perform much better than the state-of-the-art ZSL models
in most cases, which turns out to also be the case in captioning.

For qualitative examination, we present 
visual output examples in Figure~\ref{fig:zscresults}, along with the corresponding GZSD detection results. It can be observed that
the ZSC model is able to generate semantically sound captions in a variety of challenging scenes involving
both seen and unseen class instances.

\subsection{V-METEOR experiments}
\label{sec:v-meteor}

We now evaluate the baseline and proposed models using the V-METEOR metric. We present
the overall average V-METEOR scores in Table~\ref{table:v-meteor}.  These summary results show that
the proposed approach greatly improves the visual captioning score from $0.0$ to $12.63$ and also
increases the non-visual V-METEOR scores from $20.50$ to $22.26$. The final V-METEOR score improves
from $0.0$ to $13.19$. These results show that the integration of an (accurate) GZSD can not only
help with visual coverage of the captioning results but also improve the non-visual parts of the
generated captions thanks to the better visual information from the detector to the language model.
In these results, we also observe the main advantage of the proposed V-METEOR metric by being able
to separately discuss the visual and non-visual quality of the generated captions.

\begin{table}[]
\centering
{\small %
\begin{tabular}{cccc}
\toprule
Method       & V-METEOR\textsubscript{vis} & \multicolumn{1}{l}{V-METEOR\textsubscript{nvis}} & \multicolumn{1}{l}{V-METEOR} \\ 
\midrule
NBT-Baseline & 0.0         & 20.50                                 & 0.0                        \\ 
Our Method   & 12.63        & 22.26                                 & 13.19    \\
\bottomrule
\end{tabular}}
\caption{V-METEOR comparison results. V-METEOR\textsubscript{vis} represents a sub-metric that only includes results for visual words, and V-METEOR\textsubscript{nvis} represents an another sub-metric that only includes non-visual words.}
\label{table:v-meteor}
\end{table}

\begin{figure*}
\begin{center}
\includegraphics[width=\textwidth]{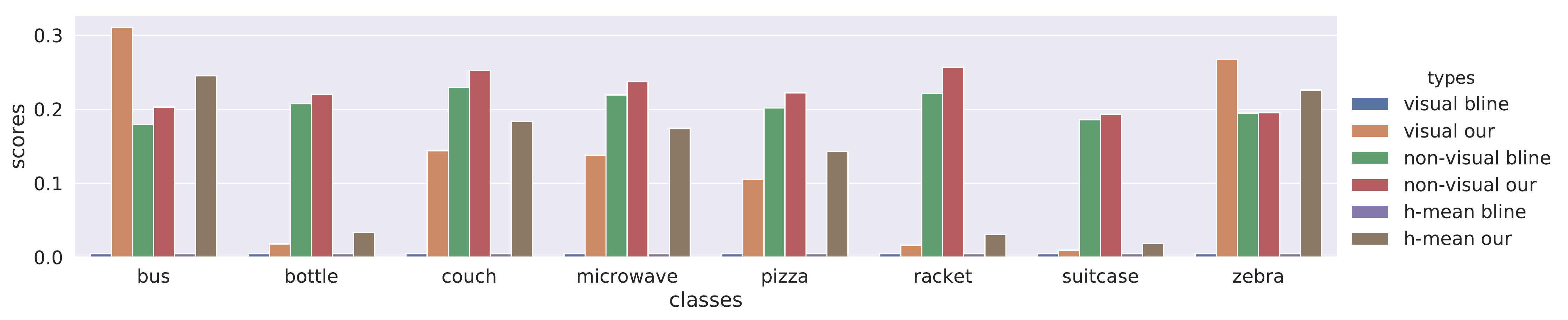}
\end{center}
\caption{V-METEOR results of each unseen classes. 
   \textbf{visual-bs} represents the visual meteor scores of the NBT-Baseline, \textbf{non-visual-bs} represents the non-visual meteor scores of the NBT-Baseline and \textbf{hm-bs} represents the V-METEOR scores of the NBT-Baseline method. Similarly, \textbf{visual, non-visual} and \textbf{hm} bars correspond to our method. (Best viewed in color.)}
\label{fig:vmeteor-detail}
\end{figure*}

\begin{figure*}
\centering
\resizebox{.85\textwidth}{!}{%    
\begin{tabular}{ccc}
\HangBox{\includegraphics[width=4.2cm, height=3.4cm]{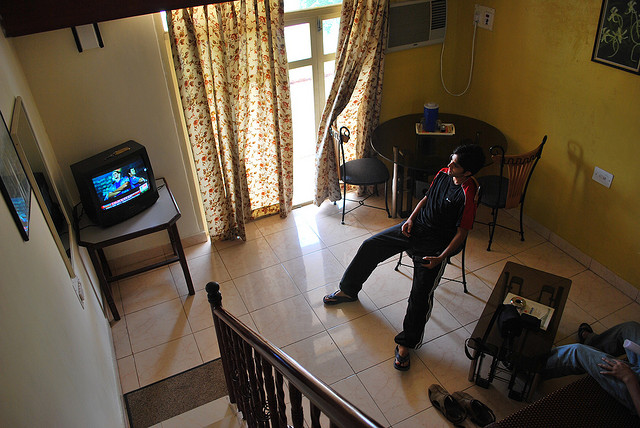}\\
$\blacklozenge$: A couple of \textbf{people} that\\ are in a room.\\
$\bigstar$: A \textbf{person} sitting in a\\ \textbf{couch} in a room.
}
\HangBox{\includegraphics[width=4.2cm, height=3.4cm]{images/captioning/COCO_val2014_000000119402.jpg}\\
$\blacklozenge$: A yellow and black \textbf{train}\\ traveling down the road.\\
$\bigstar$: A yellow and black \textbf{bus}\\ driving down a road.
}
\HangBox{\includegraphics[width=4.2cm, height=3.4cm]{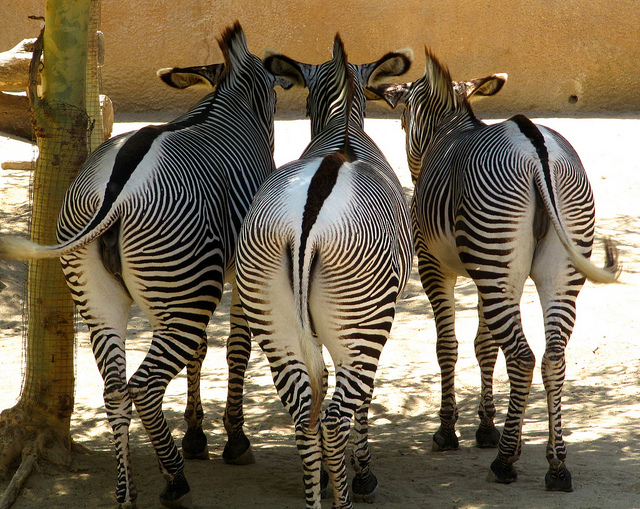}\\
$\blacklozenge$: A couple of \textbf{elephants}\\ standing next to each other.\\
$\bigstar$: A couple of \textbf{zebra}\\ standing next to each other.}
\end{tabular}}
\\
\caption{Image captioning results of NBT-baseline and our methods. $\blacklozenge$ represents the NBT-baseline results, and $\bigstar$ represents the results of the proposed method. \textbf{Bold} type words represent visual words from detectors.}
\label{fig:nbt_baseline_vs_ours}
\end{figure*}

To better understand the captioning results, we present per-class V-METEOR scores for the unseen
classes in Figure~\ref{fig:vmeteor-detail}.  In these results, we again observe both the most
significant improvements are in $\text{V-METEOR}_\text{vis}$ scores with still noticeable
improvements in non-visual scores. The complementary qualitative captioning comparisons presented
in Figure~\ref{fig:nbt_baseline_vs_ours} supports these quantitative observations: in the {\em
person} and {\em bus} examples, the whole sentence changes and improves with the correction in
visual details. In the {\em bus} and {\em zebra} examples, we observe that the NBT-baseline method
produces coarsely plausible sentences, however, with incorrect visual coverage due to confusions
across visually similar classes.

\subsection{Additional analyses}
\label{sec:analyses}

In this section, we present a quantitative analysis on the error patterns and an ablative study on
the importance of proposed similarity embeddings in GZSD.

\subsubsection{Diagnosing errors}

The experimental results show that GZSD plays a central role in achieving accurate captioning
results.  Therefore, it is potentially valuable to understand the typical detection errors of our
GZSD model, towards building better GZSD and ZSC approaches.  For this purpose, we embrace the
detector analysis approach by Hoiem \textit{et al.}~\cite{hoiem2012diagnosing}, which is originally
proposed for analyzing false positives in supervised detectors.  The original analysis approach
defines semantic categories for the PASCAL VOC dataset.  To utilize this technique in the GZSD
setting, we use the MS-COCO superclasses, namely \textit{vehicle, outdoor, animal, accessory,
sports, kitchen, food, furniture, electronic, appliance} and {\em indoor}, as defined in
\cite{lin2014microsoft}. Following \cite{hoiem2012diagnosing}, we additionally define a separate
singleton superclass for the {\em person} class, as it contains a greatly larger number of
instances and its overall distinct visual characteristics.

\begin{figure*}
\begin{center}
    \includegraphics[width=\textwidth]{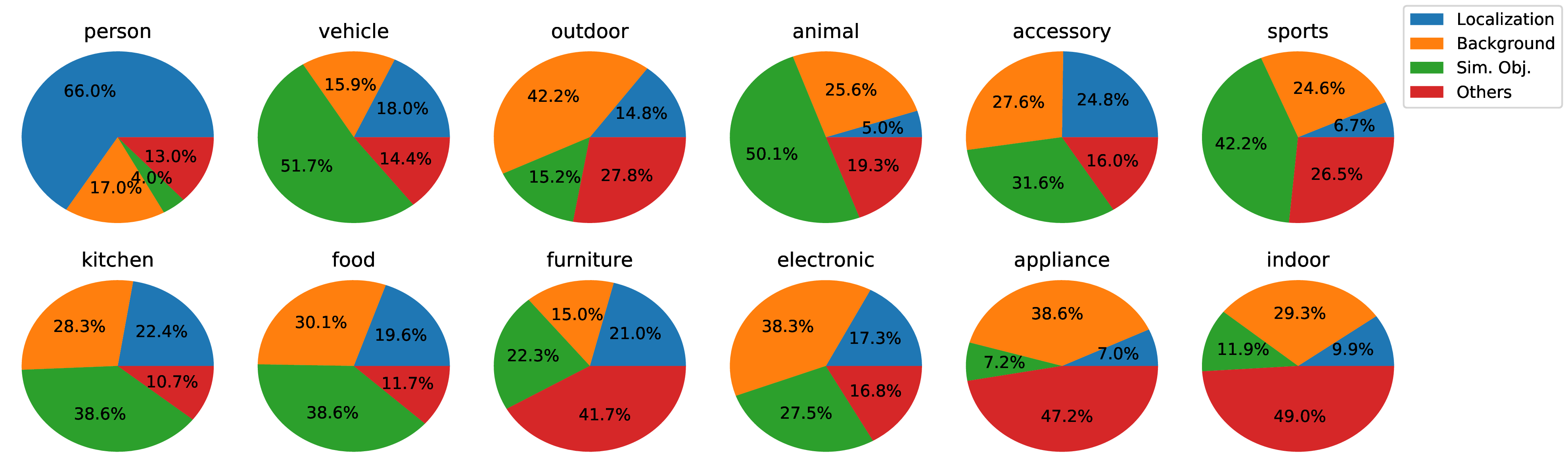}
\end{center}
\caption{False positive analyses for superclasses on MS-COCO. \textbf{Localization} represents
    detections considered as false positive due to poor localization, \textbf{Background} represents
    false positive detections located in the background, \textbf{Sim. Obj.} stands for
    misclassifications within superclass members and \textbf{Others} stands for confusion with
    other classes. (Best viewed in color.)} \label{err_analysis_super_classes}
\end{figure*}

The following four misdetection categories are examined for each superclass: (i) localization
errors, corresponding to  detections considered as false positive due to poor localization, (ii)
confusion with background, counting false positive detections located in the background, (iii) class
confusion within superclass members, and (iv) class confusion across superclasses. The
corresponding error distributions are shown in Figure~\ref{err_analysis_super_classes}.  

The obtained error distribution results show that the false positives are mainly occurred due to the
within superclass confusions for the {\em vehicle, animal, accessory, sports, kitchen} and {\em
food} superclasses. The dominant misdetection type for the {\em furniture, appliance} and {\em
indoor} superclasses is confusion with other classes. In contrast, most {\em person} misdetections
correspond to localization errors. Finally, we observe that most problematic detections for {\em
outdoor} and {\em electronic} superclasses correspond to background detections.
Overall, these results show that there is no single error pattern dominating the GZSD outputs, and
errors vary greatly across the classes.

\subsubsection{Impact of using similarity embeddings} 

One of the advantages of using the proposed class-to-class similarity vectors is that each dimension
of the embedding explicitly corresponds to a class relevance value. We additionally utilize its
structure in the design of our alpha scaling training scheme. To better understand the value of the
proposed class embeddings for GZSD, we present a direct comparison
between using the proposed class embeddings versus the original class name word embeddings.

We present the results based on both embeddings in Table~\ref{table:gzsd_results_vw_directly}.  
The results show that the 
standard word embedding scheme obtains $28.41\%$ mAP on seen classes, $14.36\%$ mAP on unseen
classes and a harmonic mean score of $19.08$. In contrast, the proposed embedding yields $28.91\%$
$15.78\%$ and $20.41$ unseen mAP, seen mAP and harmonic mean scores, respectively.
These results
show that using class-to-class similarity vectors also provides a relative performance advantage in
terms of model performance, while also enabling our effective alpha coefficient
learning procedure.

\begin{table}[]
\centering
\begin{tabular}{ccccc}
\toprule
Method & \multicolumn{1}{l}{seen/unseen} & seen  & unseen & HM\\
\midrule
Word embeddings     & 65/15                           & 28.41 & 14.36  & 19.08\\
SimEmb              & 65/15                           & 28.91 & 15.78  & 20.41\\
\bottomrule
\end{tabular}
\caption{mAP results on MS-COCO dataset with GZSD (65/15) settings, using
    the word embeddings directly versus
    class-to-class similarities as class embeddings. \label{table:gzsd_results_vw_directly}}
\end{table}

\section{Conclusion}
\label{sec:conclusion}

An important shortcoming of current image captioning methods that aim training through non-paired
datasets is that they do not work in a fully ZSL setting.  These methods generate
captions for images which consist of classes not seen in captioning datasets, but they assume that
there is a ready-to-use fully supervised visual recognition model. To this end, we define the
ZSC problem, propose a novel GZSD model
and a ZSC approach based on it. We additionally introduce a practical class embedding scheme, a 
technique to improve GZSD performance via score scaling, and a novel evaluation method that
provides insights into the ZSC results. Our qualitative and quantitative
experimental results show that our method yields promising results towards achieving our
ZSC goals.  We believe that ZSC is an important research direction
towards building captioning models that are more suitable to use in realistic, in-the-wild settings.

\section*{Acknowledgements}

This work was supported in part by the TUBITAK Grants 116E445 and 119E597. The numerical calculations reported in this paper were partially performed at TUBITAK ULAKBIM, High Performance and Grid Computing Center (TRUBA resources).

\bibliography{IEEEabrv,egbib}

\end{document}